\documentclass{bmvc2k}

\title{\metabbr : A Training-Free Approach for Handling Loosely Aligned Visual Conditions in ControlNet}

\addauthor{Woosung Joung*}{mung3477@korea.ac.kr}{1}
\addauthor{Daewon Chae*}{daewon@umich.edu}{2}
\addauthor{Jinkyu Kim}{jinkyukim@korea.ac.kr}{1}

\addinstitution{
 Department of Computer Science and Engineering\\
 Korea University\\
 Seoul, Republic of Korea
}
\addinstitution{
 Electrical and Computer Engineering\\
 University of Michigan\\
 Ann Arbor, MI, USA
}

\newcommand{\metabbr}{SemanticControl\xspace}
\newcommand{\myparagraph}[1]{\vspace{5pt}\noindent{\bf #1}}
\usepackage{cleveref}
\usepackage{enumitem}
\usepackage{wrapfig}
\usepackage[table]{xcolor} 

\usepackage{multirow}
\usepackage{booktabs}

\runninghead{Woosung Joung*, Daewon Chae*, Jinkyu Kim}{\metabbr}

\begin{document}

\maketitle

\vspace{-1.5em}
\begin{abstract}
ControlNet has enabled detailed spatial control in text-to-image diffusion models by incorporating additional visual conditions such as depth or edge maps.
However, its effectiveness heavily depends on the availability of visual conditions that are precisely aligned with the generation goal specified by text prompt—a requirement that often fails in practice, especially for uncommon or imaginative scenes. 
For example, generating an image of a cat cooking in a specific pose may be infeasible due to the lack of suitable visual conditions.
In contrast, structurally similar cues can often be found in more common settings—for instance, poses of humans cooking are widely available and can serve as rough visual guides.
Unfortunately, existing ControlNet models struggle to use such loosely aligned visual conditions, often resulting in low text fidelity or visual artifacts. 
To address this limitation, we propose \metabbr, a \textit{training-free} method for effectively leveraging misaligned but semantically relevant visual conditions.
Our approach adaptively suppresses the influence of the visual condition where it conflicts with the prompt, while strengthening guidance from the text.
The key idea is to first run an auxiliary denoising process using a surrogate prompt aligned with the visual condition (e.g., ``a human playing guitar'' for a human pose condition) to extract informative attention masks, and then utilize these masks during the denoising of the actual target prompt (e.g., ``cat playing guitar'').
Experimental results demonstrate that our method improves performance under loosely aligned conditions across various conditions, including depth maps, edge maps, and human skeletons, outperforming existing baselines. Our code is available at \url{https://mung3477.github.io/semantic-control}.

\end{abstract}

\section{Introduction}
\begin{figure}[!t]
  \centering
  \includegraphics[width=.9\linewidth]{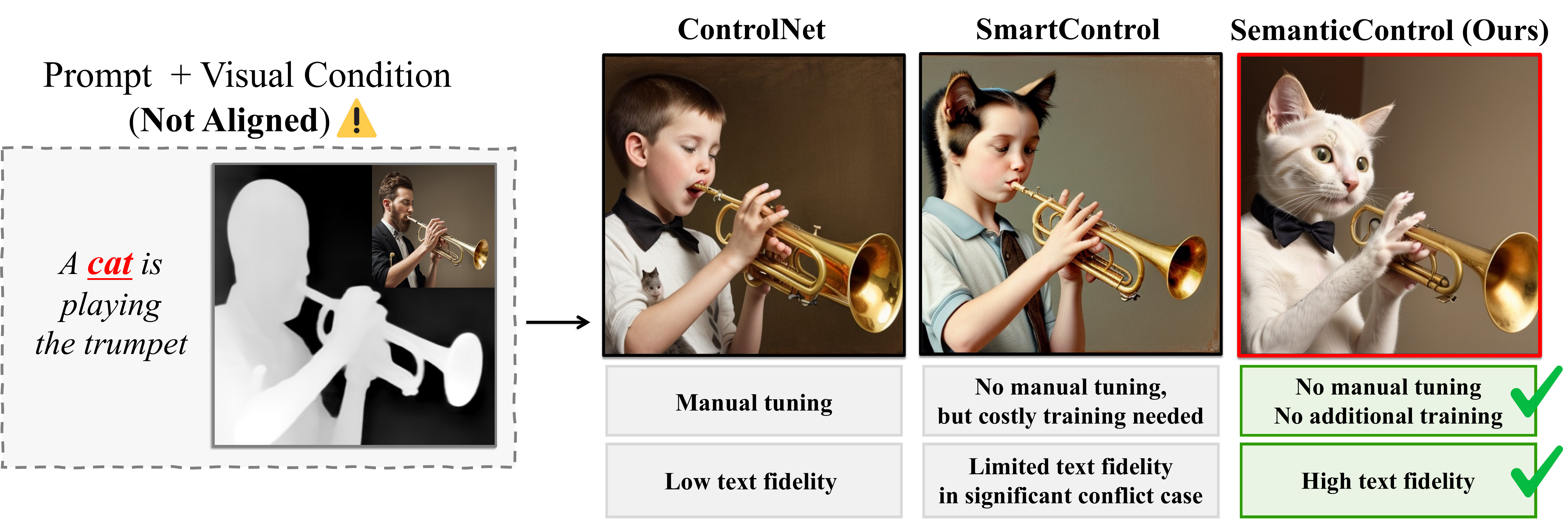}
  \caption{Comparison with prior approaches. Our method achieves high text fidelity under significant prompt–condition mismatch without manual tuning or additional training.}
  \label{fig:teaser}
  \vspace{-1em}
\end{figure}

Text-to-image diffusion models~\cite{stablediffusion,imagen, sdxl} have demonstrated remarkable capabilities in generating high-quality images from natural language prompts. However, relying solely on textual description often limits the model’s ability to accurately control spatial composition within the generated images. To address this limitation, recent methods such as ControlNet~\cite{ControlNet} and IP-Adapter~\cite{Ip_adapter} incorporate additional visual conditions (e.g., depth maps, edge maps) to provide more precise spatial guidance during generation. 

While these methods provide improved spatial controllability, their effectiveness is fundamentally constrained by the need for visual conditions that are precisely aligned with the semantics of the target prompt. In practice, obtaining such aligned conditions is often infeasible—particularly when generating uncommon or imaginative scenarios. For instance, generating an image of a cat playing a trumpet in a specific pose requires a matching visual condition that is unlikely to exist. In such cases, a more practical alternative is to use a structurally similar pose—such as that of a human in the same action—but existing methods often fail to leverage such rough conditions, producing images with low text fidelity or noticeable visual artifacts (see Figure~\ref{fig:teaser}). This inability to handle loosely aligned but semantically relevant visual conditions significantly limits the practical applicability of current spatial guidance methods.

One notable approach to tackle this issue is SmartControl~\cite{smartcontrol}, which mitigates prompt –condition misalignment in ControlNet by adaptively reducing the control intensity of the visual condition. Specifically, it introduces a predictive module trained to estimate suitable control strengths during inference. However, training this module requires a curated set of successful generations under misaligned prompt–condition pairs, which are difficult to obtain in practice. To construct such data, SmartControl manually varies control intensity across prompts and selects outputs that preserve text fidelity—requiring extensive human effort. Moreover, since ControlNet models are typically trained separately for each type of visual condition, the predictive module must also be trained per variant, limiting generalizability. Most importantly, SmartControl often fails when the mismatch is more substantial, such as when the prompt and condition involve entirely different subjects (e.g., human vs. cat). This is not only because successful training examples are scarce under such severe mismatches, but also because the method focuses solely on suppressing the visual condition, without a complementary strategy to strengthen guidance from the text prompt.

\begin{figure*}[!t]
  \centering
  \includegraphics[width=1.0\linewidth]{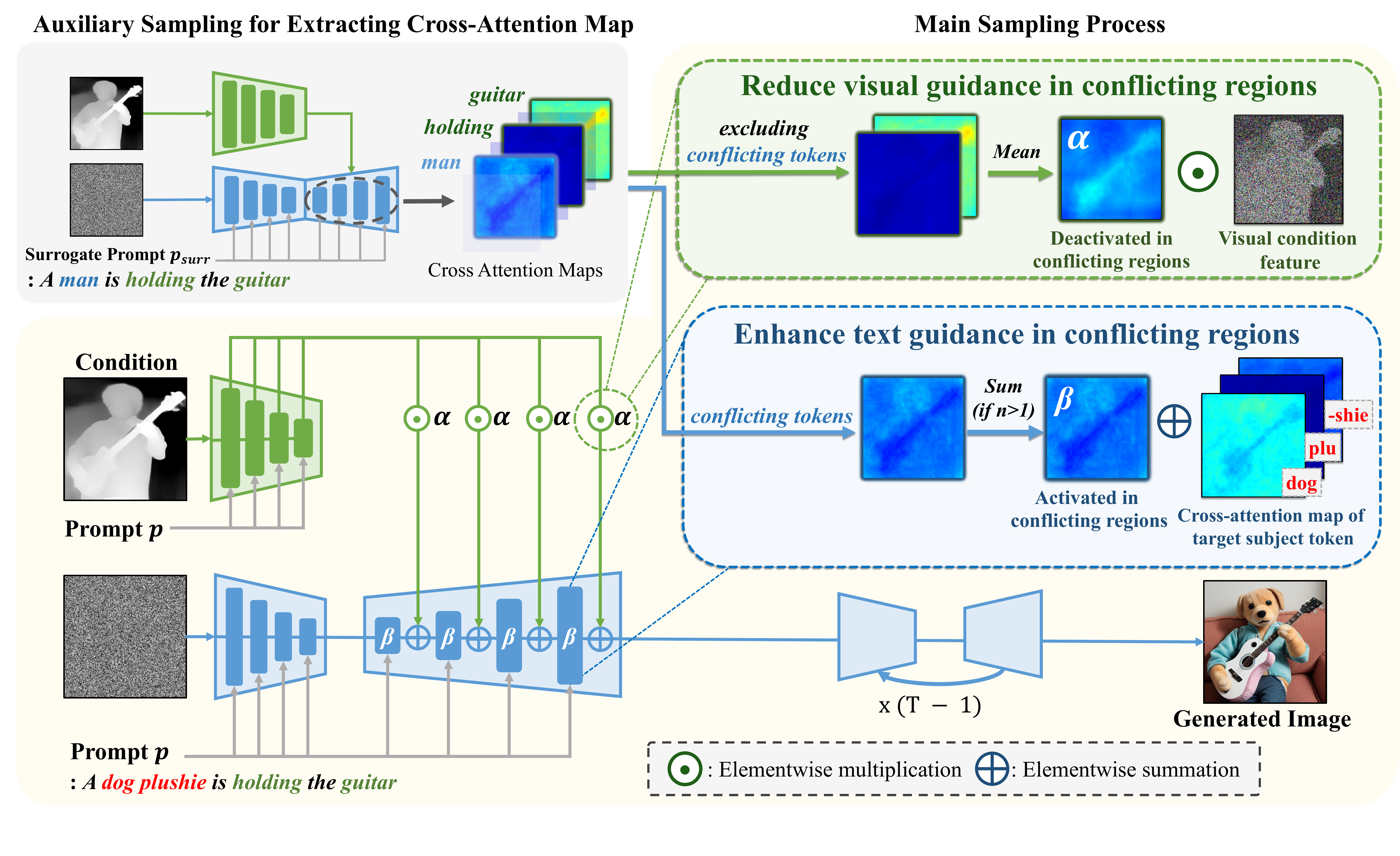}
  \caption{
    Overall pipeline of \metabbr. Using a visual condition and a surrogate prompt that describes it, we first prepare cross-attention maps of \textit{non-conflicting tokens} (``holding'', ``guitar'').  We infer a proper control scale mask $\mathbf{\alpha}_l$ for each layer by averaging those maps. We also calculate a cross-attention bias $\mathbf{\beta}$ by summing up the attention weights of \textit{conflicting tokens} (``man''). During generation, ControlNet features are multiplied with $\mathbf{\alpha}$ and cross-attention weights of \textit{target tokens} (``dog plushie'') are added by $\mathbf{\beta}$.}
  \label{fig:pipeline}
  \vspace{-1.0em}
\end{figure*}

In this work, we propose \metabbr, a \textit{training-free} enhancement to ControlNet that enables robust generation under loosely aligned visual conditions. The core idea is to suppress misleading visual signals while compensating for weakened text guidance in the conflict region. Given a target prompt and misaligned visual condition (e.g., “a cat playing guitar” with a depth map of a man playing the guitar), we first derive a surrogate prompt (e.g., “a man playing the guitar”) that is semantically aligned with the condition and perform a auxiliary denoising step to extract token-wise cross-attention maps. From these maps, we compute a suppression mask by aggregating attention from non-conflicting tokens (e.g., “playing,” “guitar”) and apply it to modulate the control intensity of the visual condition, suppressing guidance in irrelevant regions while preserving it where the semantics align. At the same time, the attention map of the conflicting token (e.g., “man”) is incorporated into the cross-attention of the corresponding target token (e.g., “cat”) during denoising with the target prompt. This prevents the target subject from being suppressed by residual visual signals entangled within the noise, ensuring that it is accurately reflected in the generated image (see Figure~\ref{fig:pipeline}). As a result, via selective attention maps based on surrogate prompts, \metabbr improves generation quality in cases where the prompt and visual condition are not fully aligned, without requiring retraining or curated data. Experimental results show that our method outperforms existing baselines under loosely aligned conditions across various inputs—including depth maps, edge maps, and human skeletons—and is also strongly preferred in human evaluations.

\section{Related Work}
\myparagraph{Controllable Text-to-Image Generation.}
To enhance controllability in text-to-image generation, various methods incorporate additional inputs beyond text. LayoutDiffuion~\cite{layoutdiffusion} controls the layout of the image via bounding boxes of objects. T2I-Adapter~\cite{t2iadapter_mou2023t2iadapterlearningadaptersdig} and ControlNet~\cite{ControlNet} guide generation with structural visual conditions (e.g., depth maps) by attaching auxiliary networks to Stable Diffusion~\cite{stablediffusion}. Several extensions of ControlNet further enable multi-condition control with a single model~\cite{unicontrol, unicontrolnet, cocktail}. 
Nevertheless, earlier studies have demonstrated that ControlNet may exhibit contention between the frozen Stable Diffusion backbone and its auxiliary condition branch \cite{HumanSD, visconet}. Especially when the visual condition conflicts with the text prompt, these models tend to favor the condition, often generating images that fail to reflect the intended semantics \cite{smartcontrol}.
SmartControl~\cite{smartcontrol} mitigates such conflicts by reducing the influence of the visual condition in regions where misalignment occurs, using a learned condition mask predictor. 


\myparagraph{Cross-Attention for Compositional Control.}
Recent text-to-image models such as Stable Diffusion \cite{stablediffusion} leverage cross-attention mechanisms to guide the generation process. Beyond serving as a spatial representation, cross-attention is central to controlling the compositionality of generated images. Several studies~\cite{divide_and_bind,prompt2prompt,linguistic_binding} have manipulated cross-attention maps to achieve specific control objectives. For example, Attend-and-Excite \cite{attend_and_excite} steers the diffusion process to ensure that all subject tokens are sufficiently attended. Prompt-to-Prompt \cite{prompt2prompt} preserves the overall structure of the original image by injecting cross-attention maps from a source prompt during early generation steps.


\section{Preliminaries}
\label{sec:preliminaries}

\myparagraph{ControlNet.} ControlNet~\cite{ControlNet} extends Stable Diffusion~\cite{stablediffusion} by enabling the model to incorporate additional visual conditions such as depth maps or human poses.
It introduces an auxiliary encoder that mirrors the structure of the UNet~\cite{UNet} used in Stable Diffusion, allowing visual conditions to be integrated into the generation process. 
Specifically, the encoder outputs from ControlNet are injected into the skip connections of the UNet decoder, guiding the spatial structure of the output.
This can be formulated as:
\begin{equation}
  \mathbf{h}_{l} = D_{l}(\mathbf{h}_{l-1} + \mathbf{s}_{l-1} + \mathbf{a}_l * \mathbf{h}_{l-1}^{CN}), \quad 1 \le l \le N-1 ,
  \label{eq:ControlNet_weight_control}
\end{equation}
where \(N\) is the number of UNet decoder blocks, \(h_{l}\) is the output of the decoder block \(D_l\), \(s_{l-1}\) is the skip connection from the corresponding encoder block, and \(h^{CN}_{l-1}\) is the ControlNet feature. The influence of ControlNet can be modulated via the control scale mask \(a_l\). 

\begin{figure}[!t]
  \centering
  \includegraphics[width=0.95\linewidth]{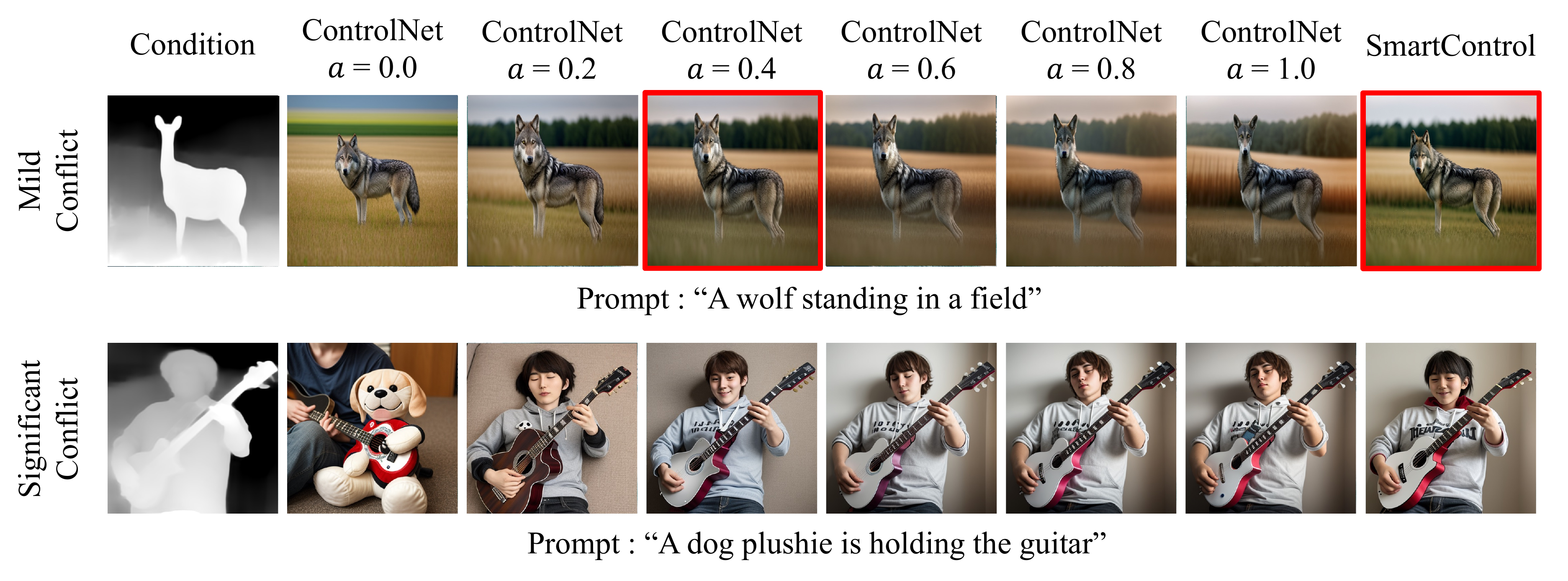}
  \vspace{-0.5em}
  \caption{
    Images were generated with different degrees of conflict between the subjects of the text prompt and the visual condition. When the conflict is mild, both ControlNet with a fixed control scale and a learned mask (SmartControl) can produce plausible results. However, under severe conflict, neither approach creates a convincing image.
  }
  \vspace{-1em}
  \label{fig:SmartControl_and_ControlNet}
\end{figure}

\myparagraph{Reducing the influence of the visual condition.} A misaligned visual condition can cause ControlNet to inject spatial features that do not fuse well with the semantics of the text prompt, often leading to the prompt being partially ignored during generation. A straightforward strategy to mitigate this issue is to reduce the influence of ControlNet features (i.e., visual guidance) by manually adjusting the control scale mask \(\mathbf{a}_l\) at \Cref{eq:ControlNet_weight_control}, as shown in the first row of \Cref{fig:SmartControl_and_ControlNet}. However, using a fixed constant control scale across all regions is often insufficient, especially when different areas of the image exhibit varying degrees of alignment or conflict. \citet{smartcontrol} highlighted this limitation and proposed a method to automatically infer a spatially varying control scale mask. They trained a mask predictor on carefully curated condition–prompt–output triplets, where each triplet was constructed using a manually selected control scale. However, collecting such training data is not only time-consuming but also fundamentally limited. As shown in the second row of \Cref{fig:SmartControl_and_ControlNet}, in cases of strong conflict—like when the prompt and condition describe entirely different subjects, there is no control scale to learn from. 

\section{Method}

\subsection{Training-Free Control Scale Estimation Using Cross-Attention}
\label{sec:ca_map_as_controlscale}

As explained in \Cref{sec:preliminaries}, ControlNet introduces conflicting spatial features when the visual condition misaligns with the text prompt, often overriding important prompt elements. Our goal is to infer a control scale mask that can filter out these conflicts and dynamically adapt to both inputs—without requiring any training. 

To achieve this, we leverage cross-attention maps from ControlNet, which reflect how much each text token contributes to specific spatial regions in the generated image. Specifically, we focus on the maps of \textbf{non-conflicting tokens}---those that express shared semantics between the prompt and condition. However, when the prompt contains conflicting semantics (e.g., “a dog plushie is holding the guitar” vs. a depth map of a man holding the guitar), ControlNet produces misaligned outputs, implying that the resulting attention maps may not meaningfully reflect how much attention each token should receive.

To address this, we use a surrogate prompt \(p_{surr}\), where conflicting tokens are replaced (e.g., “a man is holding the guitar”), aligning more closely with the visual condition. This results in a coherent output, allowing us to extract cross-attention maps that more reliably reflect the intended contribution of each token. 
We aggregate the maps of non-conflicting tokens by averaging them across decoder layers to infer the control masks. The masks will be multiplied with the ControlNet features during generation, suppressing conflicting regions in the ControlNet feature. Formally, the inferred control scale mask at layer \( l \) is defined as  
\begin{equation}
\boldsymbol{\alpha}_l = \frac{1}{N A} \sum_{n = 1}^{N} \sum_{a=1}^{A} \boldsymbol{M}_{(l,a)}[t_{n}]
\end{equation}
where \( A \) is the number of cross-attention modules in layer \( l \), N is the number of non-conflicting tokens, and \(\boldsymbol{M}_{(l,a)}[t_{n}]\) is the cross-attention map of non-conflicting token $t_n$ from the $a$th module in layer $l$.

\subsection{Cross-Attention Bias}
\label{sec:cross-attention-bias}
While attenuating ControlNet features using the estimated control scale mask helps reduce conflicts between the prompt and condition, the generated image can still fail to fully capture key elements from the prompt. For instance, when generating ``a dog plushie'', only the head may resemble a dog plushie, while the body takes on human-like features. We find that this issue often arises when \textbf{target tokens}—important tokens from \(p\) that contradict the condition—receive insufficient cross-attention. In the example ``a dog plushie is holding the guitar'', ``dog plushie'' are the target tokens that conflict with \textbf{conflicting tokens} in the surrogate prompt \(p_{surr}\), such as ``man'' in ``a man is holding the guitar''.

To address this, we introduce a cross-attention bias that amplifies the influence of target tokens during generation, inspired by \citet{zeroshot}. We compute it by reusing the cross-attention weights of the conflicting tokens from the earlier auxiliary sampling. The bias ensures that target tokens retain at least the level of influence that their corresponding tokens had during the preliminary sampling. It is defined as:
\begin{equation}
\beta_{(l, a)} = \frac{\lambda}{N_{tar}} \sum_{n = 1}^{N'} M_{(l, a)}[t_{n}]
\end{equation}
where $\lambda$ controls the strength of the bias, \(N_{tar}\) is the number of target tokens, and \(N'\) is the number of conflicting tokens. \(M_{ 
(l,a)}[t_{n}]\) denotes the cross-attention map of conflicting token \(t_n\) from the \(a\)th cross-attention module at layer \(l\). This bias is added to the cross-attention weights during generation.

\section{Experiments}
\begin{figure*}[t]
  \centering
   \includegraphics[width=1.0\linewidth]{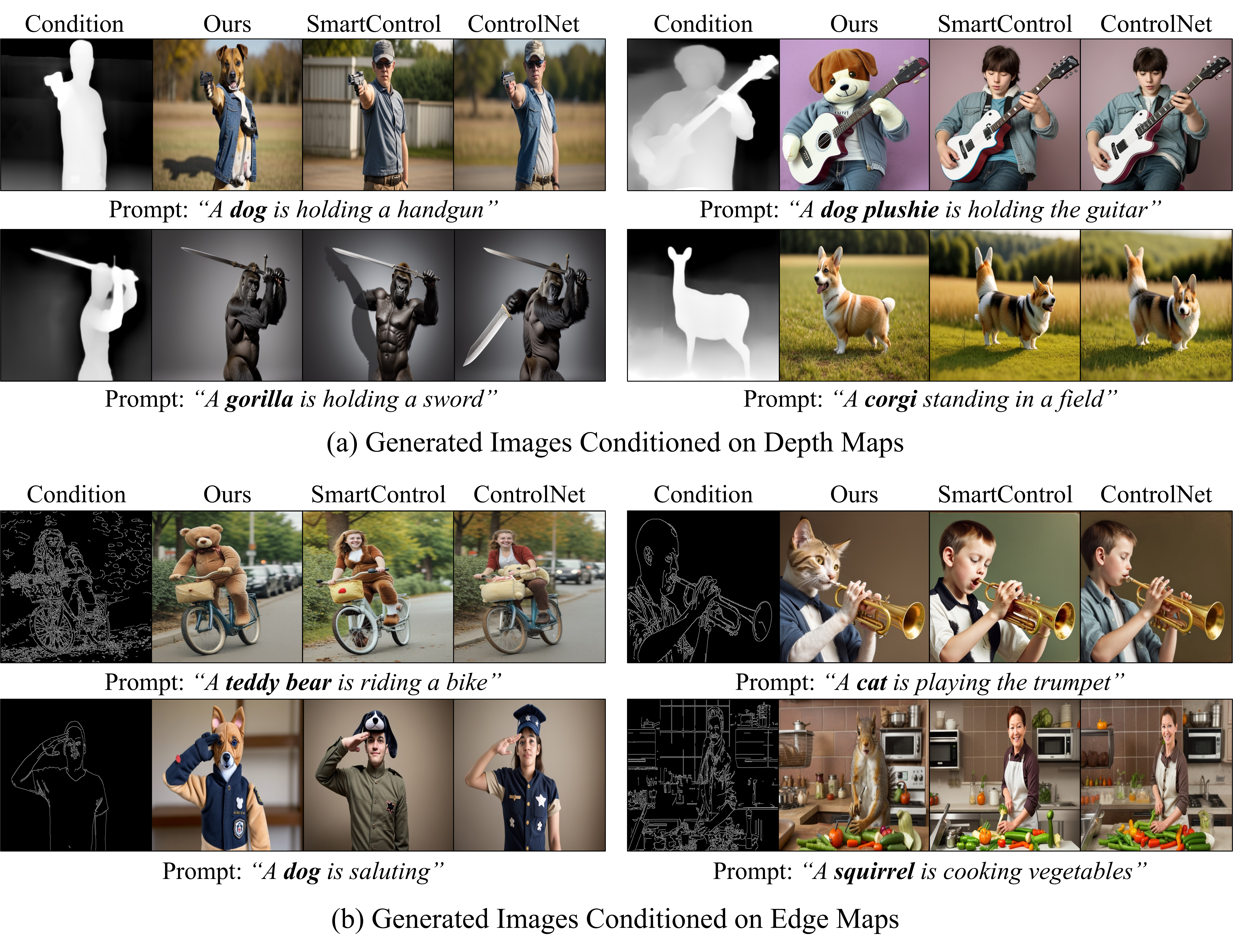}
   \vspace{-2.0em}
   \caption{
   Qualitative comparison against SmartControl and ControlNet conditioned on (a) depth and (b) edge maps. The same random seed is applied across methods for each prompt.
   }
   \vspace{-1em}
   \label{fig:qual_main}
\end{figure*}

\label{sec:exp_setups}
\myparagraph{Dataset.}
We evaluate our proposed method using samples from two sources: (i) samples obtained from the evaluation set of \citet{smartcontrol}, which consist of image-text pairs that are either well-aligned or exhibit minor misalignment between the visual content and the textual prompt; and (ii) a newly created dataset containing image-text pairs with significant semantic misalignment. To generate these latter pairs, we combine human images with prompts describing non-human subjects—for example, pairing an image of a human in a shooting pose with the prompt "a dog is shooting a gun." In total, we collect 201 such image-text pairs. More detailed explanations and examples are provided in the supplemental material.



\paragraph{Implementation and Evaluation Details.}
Our method is built upon a pretrained ControlNet architecture based on Stable Diffusion 1.5~\cite{stablediffusion}. For inference, we employ the UniPC sampler~\cite{unipc} with 50 sampling steps and a guidance scale of 7.5. Following standard evaluation protocols, we measure the alignment quality between the generated images and the corresponding text prompts using CLIP~\cite{clip} and BLIP~\cite{blip}-based similarity metrics. Additionally, we evaluate the overall quality of the generated images using ImageReward~\cite{imagereward} and PickScore~\cite{pickscore}, both of which are scoring functions trained on large-scale human preference data. Further details are provided in the supplementary material.



\begin{figure*}[t]
  \centering
   \includegraphics[width=1.0\linewidth]{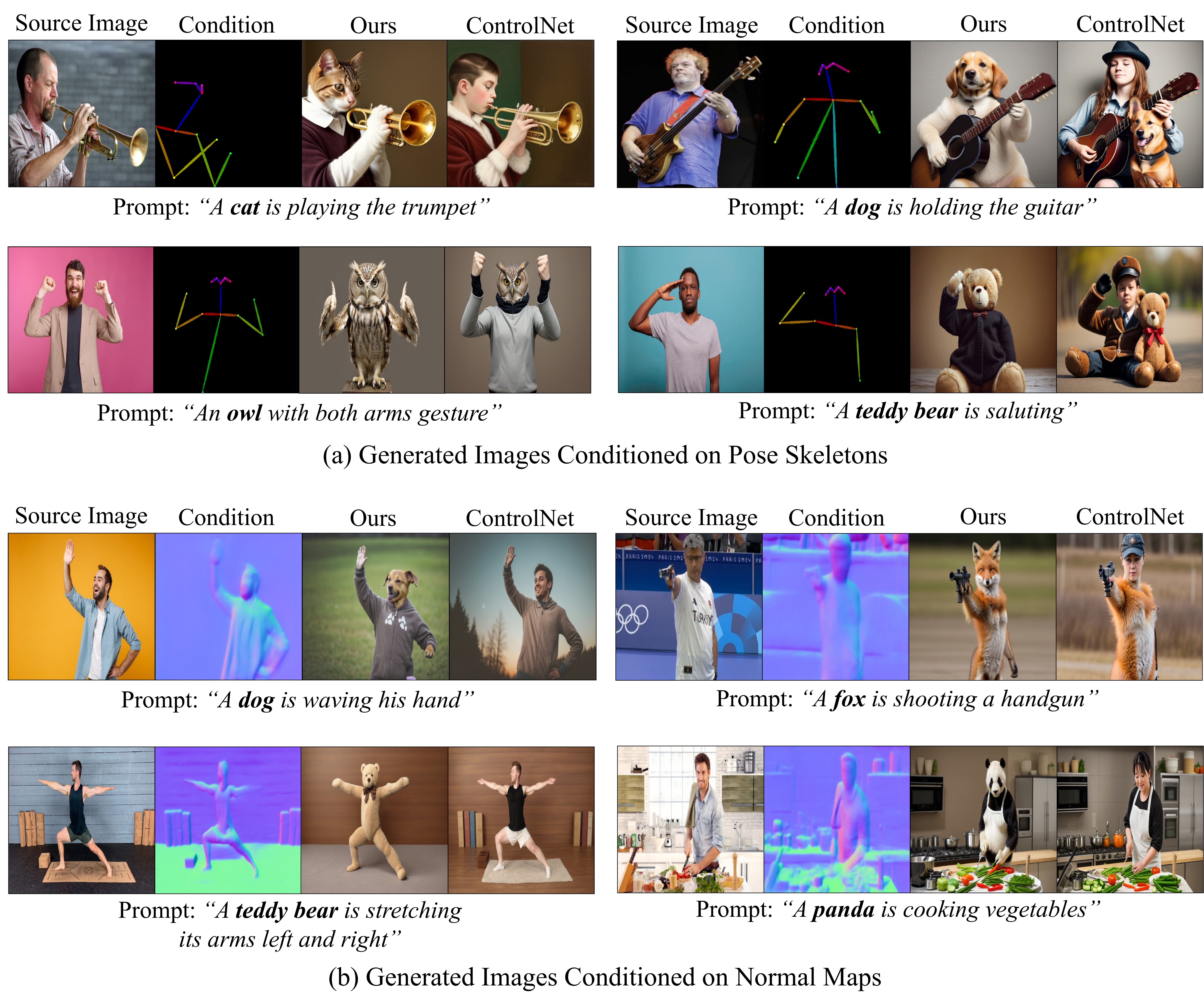}
   \vspace{-1.5em}
   \caption{Results of applying our method to ControlNet variants supporting pose skeletons and normal maps. The same random seed is applied across methods for each prompt.}
   \label{fig:diverse_condition_type}
   \vspace{-1.0em}
   
\end{figure*}
\begin{figure*}[t]
  \centering
   \includegraphics[width=1.0\linewidth]{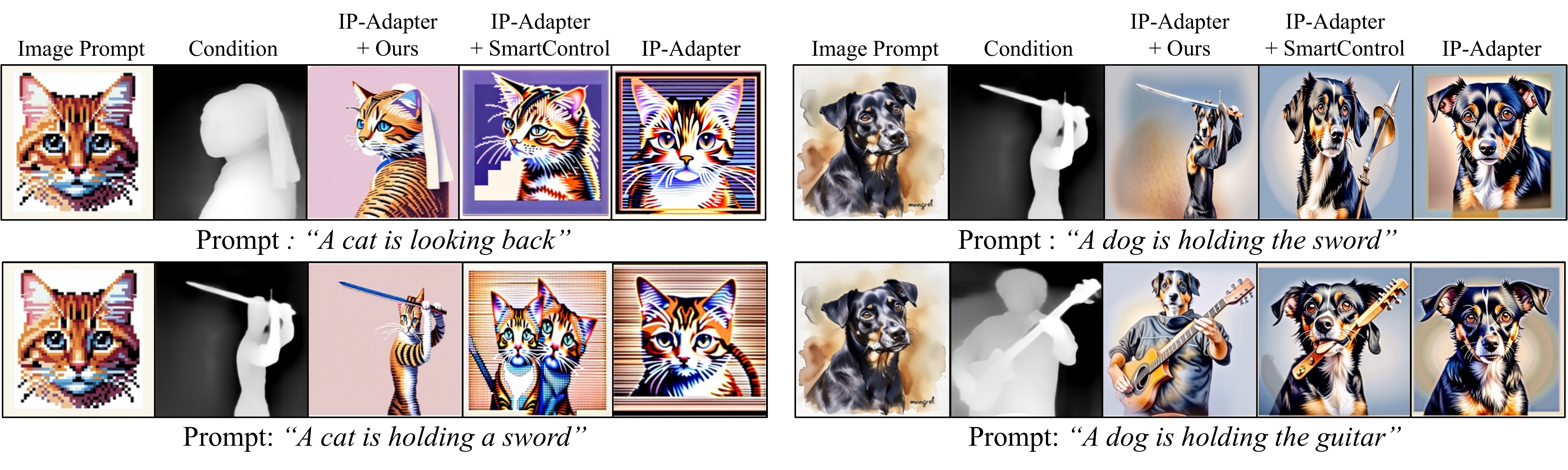}
   \vspace{-1.5em}
   \caption{Qualitative comparison between our method and SmartControl applied to IP-Adapter. The same random seed is applied across methods for each prompt.}
   \label{fig:ipadapter_result}
\end{figure*}

\myparagraph{Qualitative Analysis.}
We begin by qualitatively comparing our method with baseline models, i.e., SmartControl~\cite{smartcontrol} and ControlNet~\cite{ControlNet}\footnote{We adopt a fixed control scale of 0.4, as recommended in \citet{smartcontrol} for optimal generation quality under fixed-scale settings.}. As shown in Figure~\ref{fig:qual_main}, our approach consistently produces higher-quality images under both depth and edge conditions, demonstrating superior text fidelity. 
Notably, baseline models frequently fail to generate plausible images when the visual condition and textual prompt are misaligned, whereas our method remains robust under such scenarios. These results suggest the effectiveness of our approach in maintaining both textual fidelity and consistency with visual conditions. Additional qualitative examples are provided in the supplementary material.


\myparagraph{Analysis with Diverse Visual Conditions and Architectures.}
Since our method is entirely training-free, it can be readily applied to a wide range of ControlNet variants supporting different visual conditions. To demonstrate this, we evaluate our method on additional visual inputs such as normal maps and pose skeletons. Note that, since the official implementation of SmartControl does not supports such conditions, our comparisons are restricted to ControlNet. As shown in Figure~\ref{fig:diverse_condition_type}, our method effectively resolves misalignment between the visual condition and the text prompt under these varying visual inputs, resulting in generations with high text fidelity. This highlights the applicability of our approach across diverse control signals, without the need for condition-specific tuning.

In addition, our method is compatible with other ControlNet-based architectures. For example, we can apply our method to IP-Adapter~\cite{Ip_adapter}, which combines image prompts with pose conditions via learned adapters. As shown in Figure~\ref{fig:ipadapter_result}, our method more effectively reflects the intended text and pose conditions than existing baselines, while preserving subject fidelity to the image prompt under loosely aligned scenarios.

\begin{wrapfigure}{r}{0.41\textwidth}
    \centering
    \includegraphics[width=1.0\linewidth]{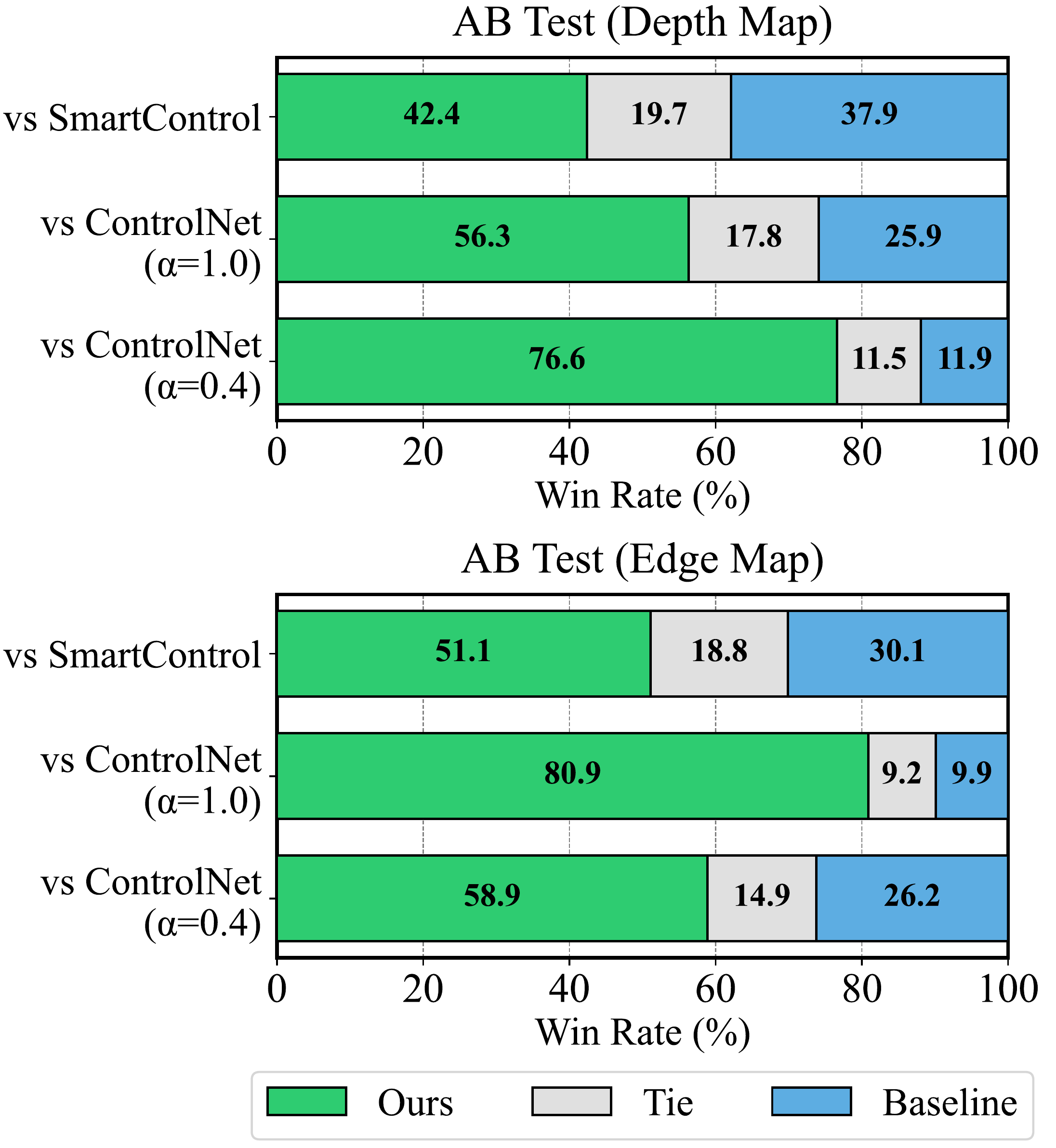}
    \caption{Human evaluation on ours and baselines. }
    \vspace{-1.0em}
    \label{fig:human_eval}
\end{wrapfigure}
\myparagraph{Human Evaluations.}
To assess the quality of images generated by our method compared to baselines, we conduct a human study. Using the 201 prompts described in the Dataset Section, we generate five images per prompt, resulting in a total of 1,005 images for each visual condition type. For each comparison, we present a pair of images—one generated by our method and the other by a baseline—alongside the corresponding visual condition and text prompt. We ask human raters to indicate their preference based on three criteria: (i) alignment with the text prompt, (ii) structural similarity to the visual condition, and (iii) overall image quality. Each query is evaluated by five independent raters on Amazon Mechanical Turk, and we aggregate the results using majority voting. 

As shown in Figure~\ref{fig:human_eval}, our method consistently outperforms the baselines under both depth and edge map conditions. Notably, in the edge map setting, our method is preferred at least 1.5 times more often than the baselines, highlighting its effectiveness in handling misaligned visual cues. We provide more details of human study in the supplementary material.

{
\setlength{\tabcolsep}{2pt}
\renewcommand{\arraystretch}{1.4} 
    \begin{table}[t]
        \begin{center}
        \caption{Quantitative comparison between our method and baselines. \textbf{Bold} indicates the best performance, and \underline{underline} indicates the second best.}
        \label{tab:performance_top5}
        \resizebox{1.0\linewidth}{!}{
        \begin{tabular}{@{}lccccccccc@{}}   
        \toprule
        \multirow{2}{*}{Methods} &\multicolumn{4}{c}{Visual Condition: Depth Map} & \multicolumn{4}{c}{Visual Condition: Canny Edge}
        \\ \cmidrule{2-5} \cmidrule{6-9}
        & CLIP($\uparrow$) & BLIP($\uparrow$) & ImageReward($\uparrow$) & PickScore($\uparrow$) & CLIP($\uparrow$) & BLIP($\uparrow$) & ImageReward($\uparrow$) & PickScore($\uparrow$) \\\midrule
        
        ControlNet ($\alpha$=1.0) & 0.2878 & 0.4283 & -0.3063 & 0.2155 & 0.2741 & 0.4042 & -0.7893 & 0.2105  \\
        ControlNet ($\alpha$=0.4)& 0.3146 & 0.4633 & 0.5744 & 0.2248 & 0.3129 & 0.4604 & 0.5488 & 0.2242  \\
        SmartControl & 0.3186 & 0.4735 & 0.8574 & 0.2271 & 0.3127 & 0.4647 & 0.5679 & 0.2246  \\
        \metabbr w/o attention bias & \underline{0.3274} & \underline{0.4813} & \underline{0.9428} & \underline{0.2291} & \underline{0.3265} & \underline{0.4797} & \underline{0.8919} & \underline{0.2285} \\
        \rowcolor{gray!20} \metabbr (Ours) & \textbf{0.3322} & \textbf{0.4904} & \textbf{1.1538} & \textbf{0.2304} & \textbf{0.3302} & \textbf{0.4864} & \textbf{1.0289} & \textbf{0.2292} \\
        \bottomrule
        \label{tab:main_quan}
        \end{tabular}
        }
        \end{center}
    \vspace{-2.0em}
    \end{table}
}

\myparagraph{Quantitative Evaluation.} 
Table~\ref{tab:main_quan} reports the results of our quantitative evaluation using the set of metrics described in the Evaluation Details Section. Our method achieves the best performance across all metrics for both types of visual conditions. These results are consistent with the qualitative analysis and human evaluation, where our method demonstrates superior image quality and strong text fidelity compared to baselines.

\begin{figure*}[t]
  \centering
   \includegraphics[width=1.0\linewidth]{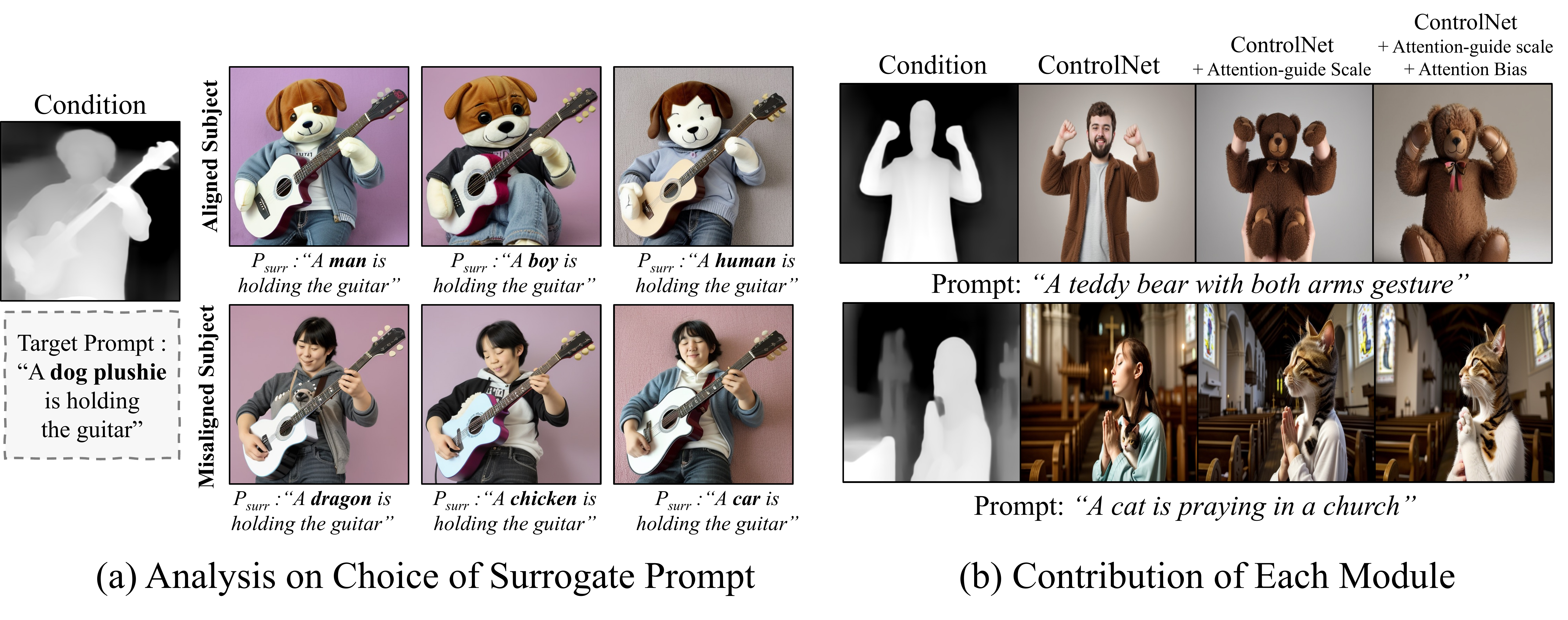}
   \vspace{-1.5em}
   \caption{Additional analysis results showing (a) the impact of surrogate prompt selection and (b) the contribution of each module to the overall performance.}
   \label{fig:ablation}
    \vspace{-0.5em}
\end{figure*}

\label{par:surrogate_ablation}
\myparagraph{Analysis on Selection of Surrogate Prompt.}
We analyze how the choice of surrogate prompt affects our method, focusing on whether it needs to be well aligned with the visual condition and whether the performance is robust to the specific selection. As our method relies on attention masks derived from the surrogate prompt, we hypothesize that the prompt must be semantically aligned with the visual condition to be effective. To verify this, we examine cases where the surrogate is not aligned with the visual input. As shown in Figure~\ref{fig:ablation} (a), using a such prompt leads to degraded performance and poor text fidelity, highlighting the need for alignment. We also test the robustness of our method to prompt choices by substituting the surrogate prompt with related terms such as “boy” and “human”. All variations produce similarly high quality results, indicating that our method is robust to prompt choices as long as they are intuitively aligned with the visual condition.

\myparagraph{Ablation Study.}
We conduct an ablation study to see the effect of each component in our method. As shown in Figure~\ref{fig:ablation} (b), the original ControlNet without our method fails to accurately reflect the subject described in the prompt, resulting in low text fidelity. When we introduce attention-guided control scaling, the subject begins to appear clearly, although some visual artifacts remain. With the additional incorporation of attention bias, these artifacts are removed, leading to high-quality generations. This qualitative observation is further supported by the quantitative results presented in Table~\ref{tab:main_quan}, which show consistent performance improvements as each component is progressively added.


\section{Conclusion and Limitation}
In this work, we propose \metabbr, a \textit{training-free} approach that enables text-to-image diffusion models to effectively utilize loosely aligned visual conditions while preserving text fidelity. By leveraging surrogate prompts, our method extracts informative cross-attention maps that modulate visual guidance based on semantic alignment with the target prompt. Through extensive experiments across various visual conditions and architectures, we demonstrate that our method consistently outperforms baselines, achieving higher human preference scores under misaligned settings. We believe that our training-free approach broadens the applicability of spatially guided generation by relaxing the requirement for precisely matched conditions and enabling robust control even in mismatched scenarios.

This work underscores the necessity of examining various forms of condition–prompt conflict. Our current approach primarily tackles a spectrum of mismatches within single-subject prompts, which represents the most common ControlNet use-case. Nonetheless, it does not yet address conflicts arising in multi-subject descriptions or spatially intricate scenes, where devising suitable surrogate prompts becomes considerably harder. Guided by the insights of this work, future research will aim to devise a systematic framework for constructing surrogate prompts capable of handling these broader categories of conflict.
\section*{Acknowledgments}
This work was supported by IITP grant funded by the Korea government(MSIT) (IITP-2025-RS-2024-00397085, 30\%, RS-2025-02263754, Human-Centric Embodied AI Agents with Autonomous Decision-Making, 30\%, IITP-2025-RS-2020-II201819, 10\%, RS-2022-II220043, Adaptive Personality for Intelligent Agents, 30\%). Daewon Chae is supported by the Hyundai Motor Chung Mong-Koo Foundation. We thank Kyungryul Back for his helpful comments and insightful discussions.

\bibliography{egbib}

\newpage
\makesupptitle
\vspace{-1em}

\setcounter{section}{0}
\section{Dataset}
\myparagraph{Mild Conflicts.} Although \citet{smartcontrol}  didn't release the evaluation dataset, we referred to their qualitative results and collected similar prompt-control pairs. We used control images of 12 subjects and various prompts that have a minor conflict with the image, including animals (dog, cat, horse, pig, deer, wolf, squirrel, hedgehog, eagle, parrot, duck), cartoon characters (hulk, spider-man, jerry, pikachu, micky mouse), fruits (apple, strawberry, pineapple, pear), vehicles (car, pick-up truck, tractor, bike, motorcycle), objects (cup, vase, candle), celebrities (obama) and landmarks (Pyramid, Arch of Triumph, Colosseum in Rome).

\myparagraph{Significant Conflicts.} We focused on significant conflicts between the subject of the text prompt and the visual condition---human and non-human. We collected 15 unique actions of humans, by gathering images from stock image services and previous works that investigated the problems related to generating human actions \cite{ADI_huang2024learningdisentangledidentifiersactioncustomized, smartcontrol}. Evaluation was conducted with 11 different non-humans (a dog, a dog plushie, a cat, a panda, a teddy bear, a polar bear, a squirrel, an elephant, a fox, a gorilla, an owl). 

\begin{figure}[t]
  \centering
   \includegraphics[width=1.0\linewidth]{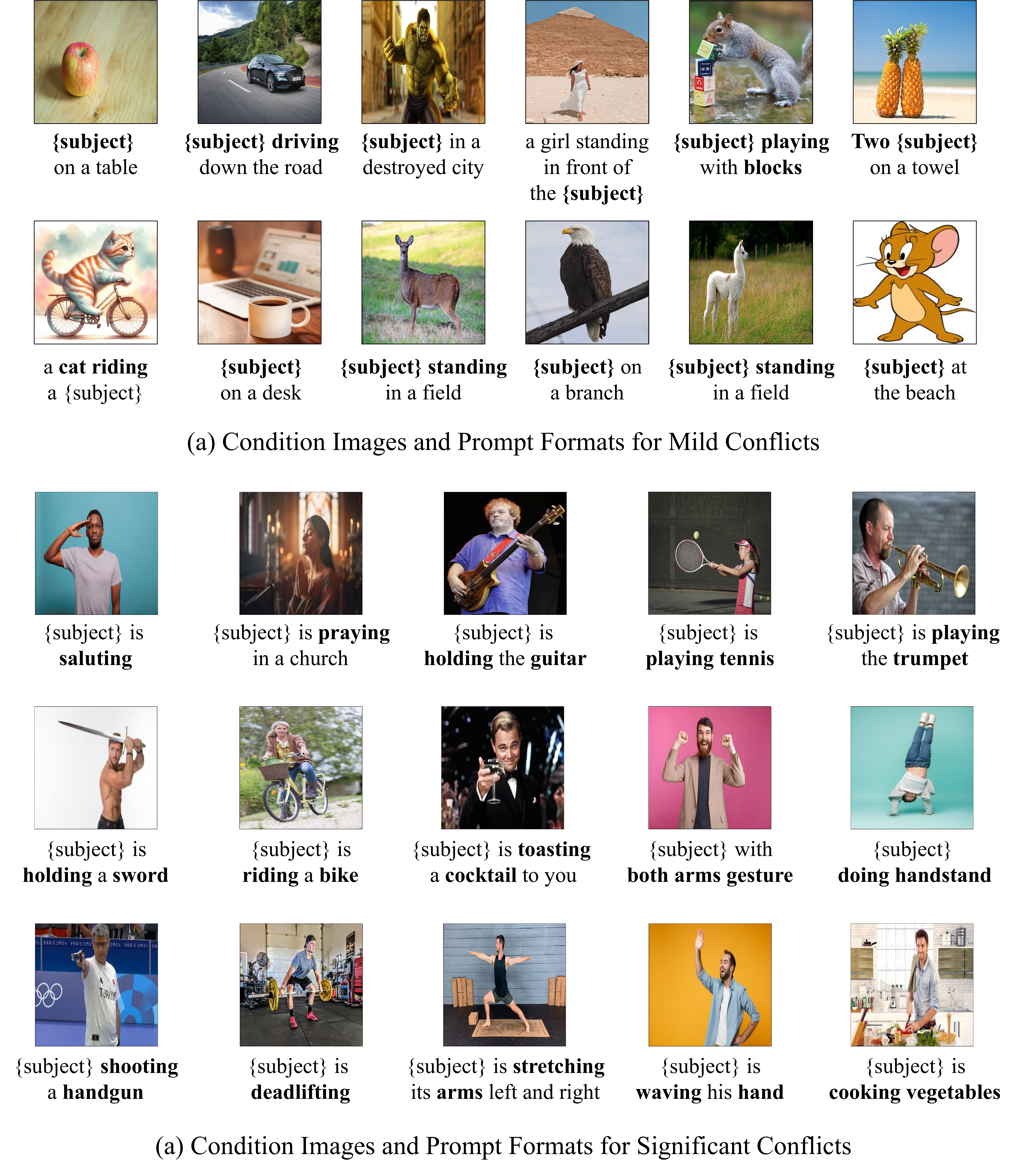}
   \vspace{-1.5em}
   \caption{
   Condition images and prompts for each level of conflict. We constructed the surrogate prompts for each condition by substituting only the subject of the prompt. \textbf{Bold} indicates non-conflicting tokens of the surrogate prompt. \(\{subject\}\) tokens are used as conflicting tokens for significant conflicts, and we did not use conflicting tokens for mild conflicts.
   }
   \vspace{-1em}
   \label{fig:supp_fig_dataset}
\end{figure}

\myparagraph{Surrogate Prompts.} Surrogate prompts should be aligned with the given visual condition. We constructed surrogate prompts for our dataset by substituting the subject of the prompt that we are trying to generate. Another key component of our method is the \textit{non-conflicting tokens}. We selected those tokens from the surrogate prompt with the following rules: (i) remove stopwords, (ii) remove tokens that are unrelated to the desired context, and (iii) exclude conflicting tokens. Examples of images and the corresponding prompts are provided on \Cref{fig:supp_fig_dataset}. 

\section{Implementation Details}

\myparagraph{Excluding Special Tokens.} We empirically found that special tokens such as \(<sot>\) and \(<eot>\) tend to receive disproportionately high attention values. Their dominance can suppress the attention assigned to other
tokens, making it difficult to assess the relative importance of meaningful semantic tokens.
Thus, before aggregating those maps to construct a control scale mask, we excluded those special tokens and re-normalized the softmax over the remaining
tokens to obtain a clearer distribution of token-level influence.

\myparagraph{Control Scale for Blocks without a Cross-attention Operator.}
Since our method is built on the pipeline of Stable diffusion \cite{stablediffusion}, the first upsampling block of the decoder in UNet does not contain cross-attention operators. Since this block also requires an inferred control scale, we empirically chose to use the control scale mask of the middle block of the UNet as the alternative. This was because both blocks have the same input image resolution.

\myparagraph{Control Bias $\lambda$.}
We use a control bias of $\lambda = 3.0$ in our main experiments. We also report results with $\lambda = 1.0$ in Table~\ref{tab:sup_quan}, where our method still outperforms SmartControl.

{
\vspace{1.0em}
\setlength{\tabcolsep}{4pt}
\renewcommand{\arraystretch}{1.2} 
    \begin{table}[h]
        \begin{center}
        \caption{Quantitative results with different control bias $\lambda$.}
        \vspace{1.0em}
        \label{tab:performance_top5}
        \resizebox{0.7\linewidth}{!}{
        \begin{tabular}{@{}lccccc@{}}   
        \toprule
        \multirow{2}{*}{Methods} &\multicolumn{4}{c}{Visual Condition: Depth Map}
        \\ \cmidrule{2-5}
        & CLIP($\uparrow$) & BLIP($\uparrow$) & ImageReward($\uparrow$) & PickScore($\uparrow$) \\\midrule
        
        SmartControl & 0.3186 & 0.4735 & 0.8574 & 0.2271   \\
        \metabbr ($\lambda$=1.0)& 0.3295 & 0.4856 & 1.0473 & 0.2298  \\
        \metabbr ($\lambda$=3.0)& 0.3322 & 0.4904 & 1.1538 & 0.2304  \\
        \bottomrule
        \label{tab:sup_quan}
        \end{tabular}
        }
        \end{center}
        \vspace{-3em}
    \end{table}
}

\section{Additional results}

\begin{figure*}[t]
  \centering
   \includegraphics[width=1.0\linewidth]{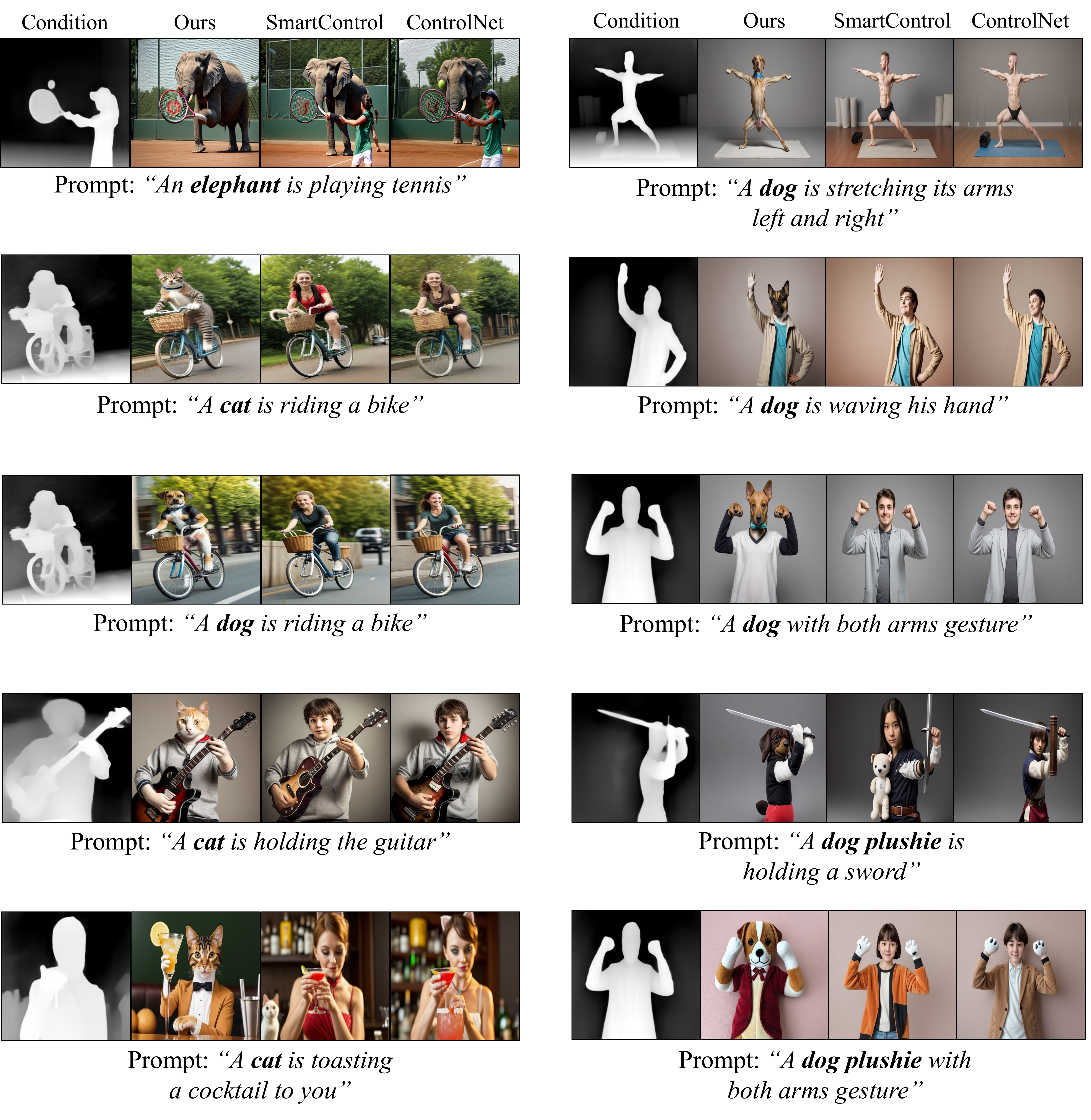}
   \vspace{-1.5em}
   \caption{
   Additional examples generated with the depth maps. The same random seed is applied across methods for each prompt.
   }
   \vspace{-1em}
   \label{fig:supp_fig_results_depth_map}
\end{figure*}

\begin{figure*}[t]
  \centering
   \includegraphics[width=1.0\linewidth]{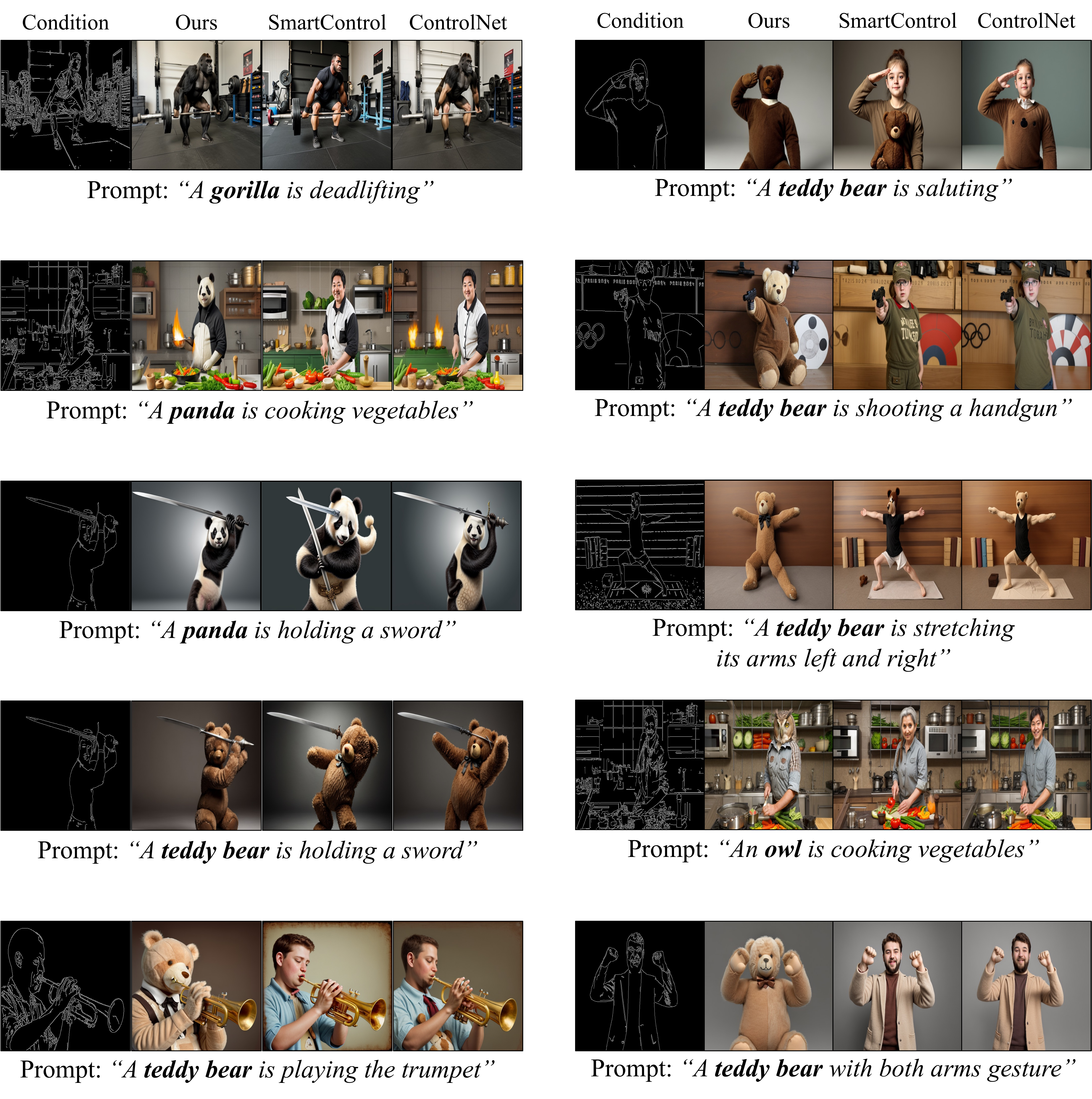}
   \vspace{-1.5em}
   \caption{
   Additional examples generated with the edge maps. The same random seed is applied across methods for each prompt.
   }
   \vspace{-1em}
   \label{fig:supp_fig_results_canny_edge}
\end{figure*}

\begin{figure*}[t]
  \centering
   \includegraphics[width=1.0\linewidth]{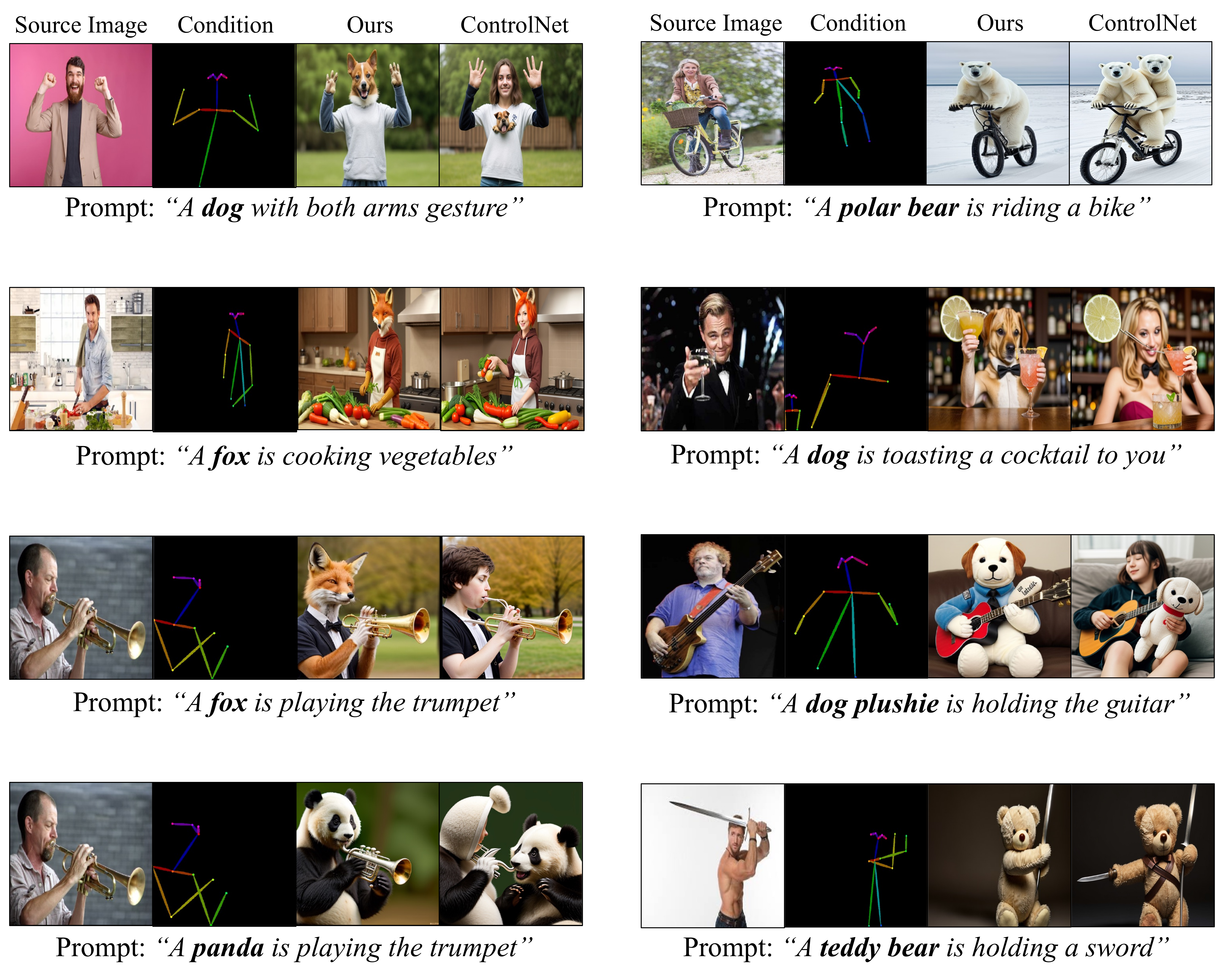}
   \vspace{-1.5em}
   \caption{
   Additional examples generated with the human skeleton of the given image. The same random seed is applied across methods for each prompt. Although generated images followed the visual condition well, detailed structures are different to the given image due to the nature of the human pose. 
   }
   \vspace{-1em}
   \label{fig:supp_fig_results_human_pose}
\end{figure*}

\begin{figure*}[t]
  \centering
   \includegraphics[width=1.0\linewidth]{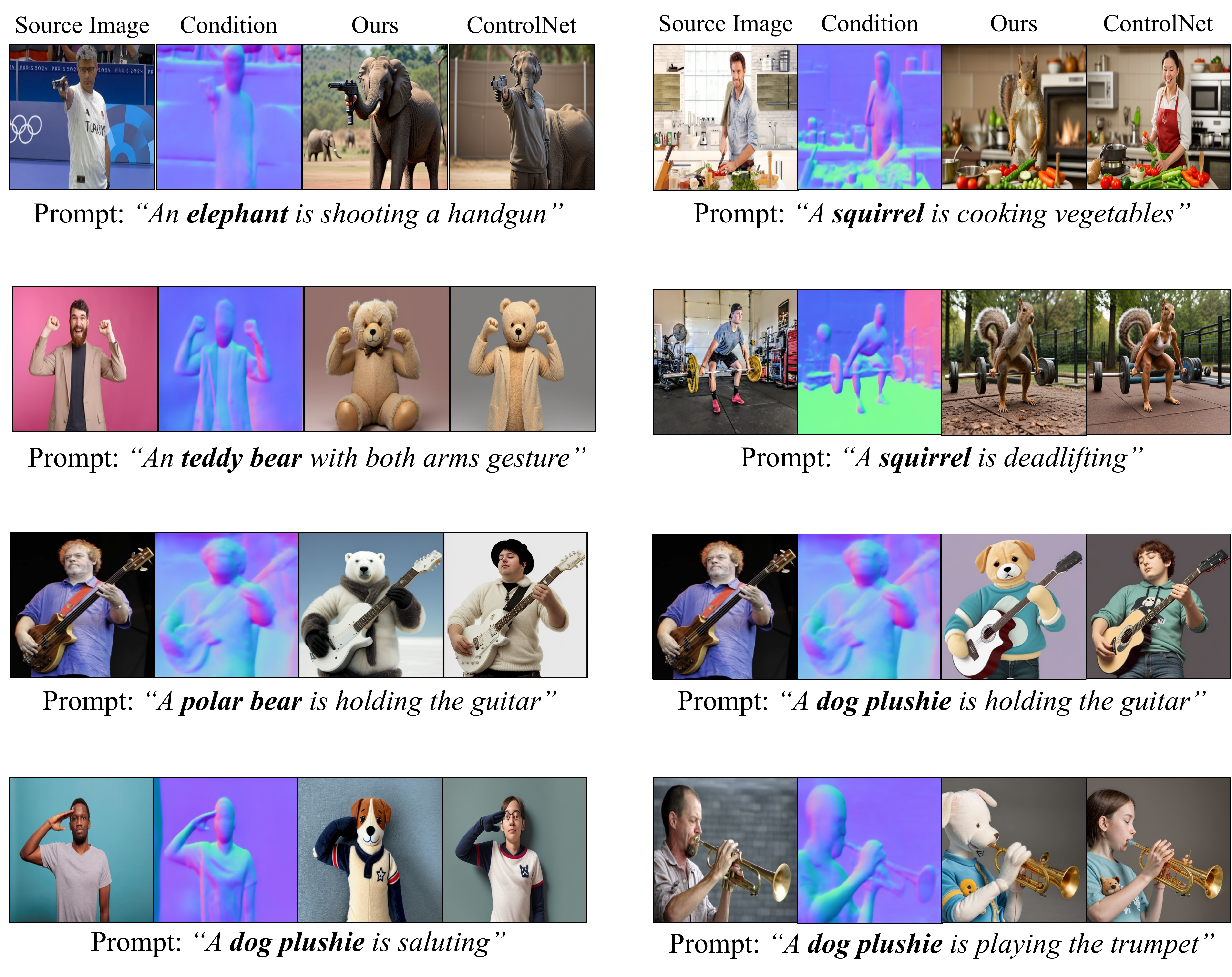}
   \vspace{-1.5em}
   \caption{
   Additional examples generated with the normal map of the given image. The same random seed is applied across methods for each prompt.
   }
   \vspace{-1em}
   \label{fig:supp_fig_results_normal_map}
\end{figure*}

We provide additional generated examples for significant conflicts using various condition types---depth maps (\Cref{fig:supp_fig_results_depth_map}), canny edges (\Cref{fig:supp_fig_results_canny_edge}), human skeletons (\Cref{fig:supp_fig_results_human_pose}), and normal maps (\Cref{fig:supp_fig_results_normal_map}). Since SmartControl does not support normal maps and human skeletons, we only compare the result with ControlNet for those cases.

\section{Details on Human evaluation}
\begin{figure*}[ht]
  \centering
   \includegraphics[width=0.85\linewidth]{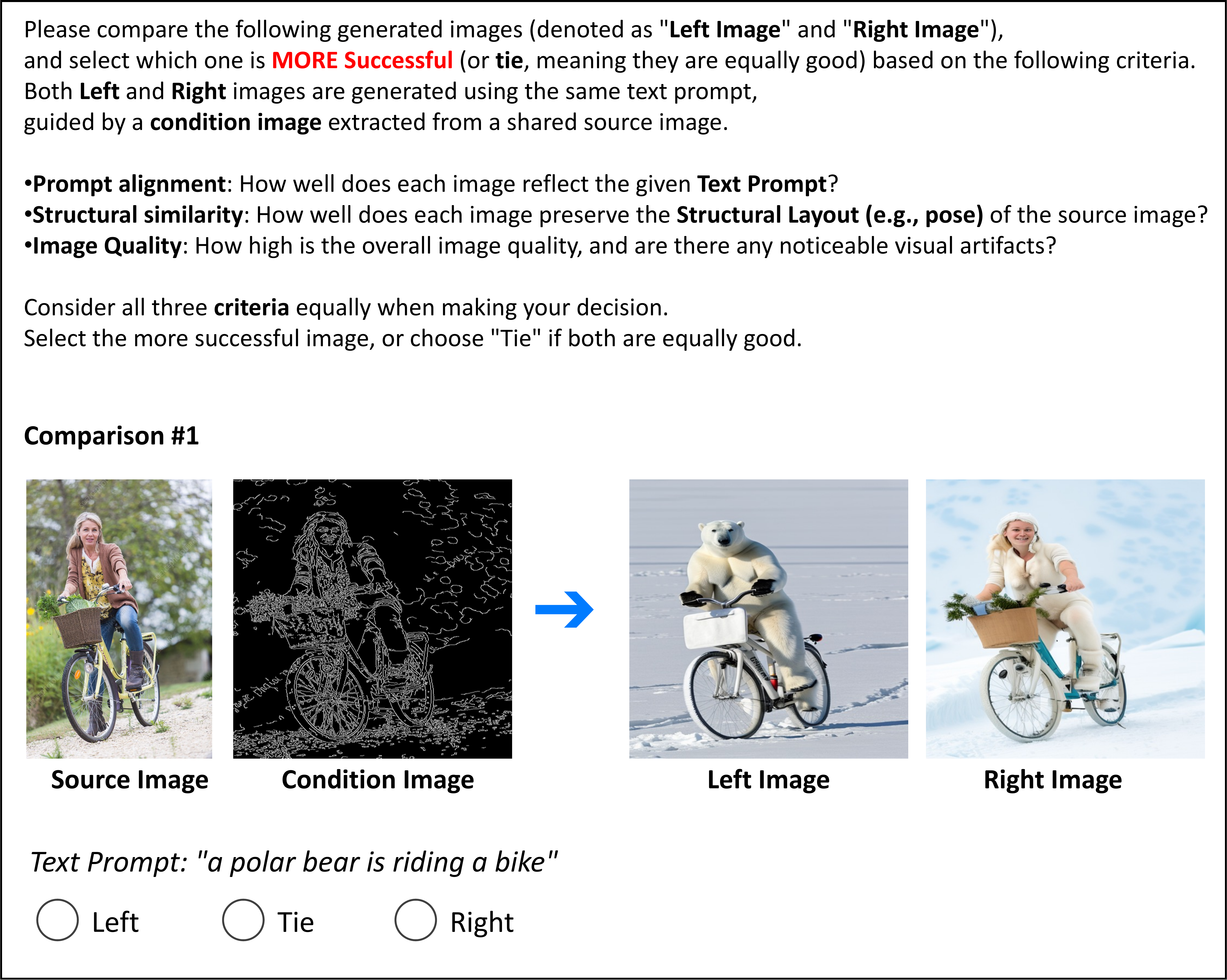}
   \caption{Human survey interface: instructions and an example questionnaire.}
   \label{fig:sup_human_eval}
\end{figure*}

As mentioned in the main paper, we conduct a human evaluation to compare our method against baseline approaches in terms of overall preference. For each visual condition, we evaluate 1,005 generated image pairs using Amazon Mechanical Turk, collecting responses from five independent raters per image. We aggregate the results through majority voting. Figure~\ref{fig:sup_human_eval} presents the actual instruction and a sample question shown to each rater.

\end{document}